\documentclass{article} 
\usepackage{iclr2023_conference,times}

\usepackage{amsmath,amsfonts,bm}

\def\eqref#1{equation~\ref{#1}}

\def\1{\bm{1}}

\DeclareMathAlphabet{\mathsfit}{\encodingdefault}{\sfdefault}{m}{sl}
\SetMathAlphabet{\mathsfit}{bold}{\encodingdefault}{\sfdefault}{bx}{n}

\newcommand{\R}{\mathbb{R}}

\usepackage{hyperref}
\usepackage{url}
\usepackage{graphicx}
\usepackage{caption}
\usepackage{subcaption}
\usepackage{wrapfig}

\title{Time-Dependent Iterative Imputation for Multivariate Longitudinal Clinical Data}

\author{Omer Noy \& Ron Shamir \\
Blavatnik School of Computer Science, Tel-Aviv University, Tel-Aviv, Israel\\
\texttt{omernoy4@gmail.com} \ \  
\texttt{rshamir@tau.ac.il}
}
\iclrfinalcopy 
\begin{document}

\maketitle

\begin{abstract}
Missing data is a major challenge in clinical research. In electronic medical records, often a large fraction of the values in laboratory tests and vital signs are missing. The missingness can lead to biased estimates and limit our ability to draw conclusions from the data. Additionally, many machine learning algorithms can only be applied to complete datasets. A common solution is data imputation, the process of filling-in the missing values. However, some of the popular imputation approaches perform poorly on clinical data. We developed a simple new approach, Time-Dependent Iterative imputation (TDI), which offers a practical solution for imputing time-series data. It addresses both multivariate and longitudinal data, by integrating forward-filling and Iterative Imputer. The integration employs a patient, variable, and observation-specific dynamic weighting strategy, based on the clinical patterns of the data, including missing rates and measurement frequency. We tested TDI on randomly masked clinical datasets. When applied to a cohort consisting of more than 500,000 patient observations from MIMIC III, our approach outperformed state-of-the-art imputation methods for 25 out of 30 clinical variables, with an overall root-mean-squared-error of 0.63, compared to 0.85 for SoftImpute, the second best method. MIMIC III and COVID-19 inpatient datasets were used to perform prediction tasks. Importantly, these tests demonstrated that TDI imputation can lead to improved risk prediction.
\end{abstract}

\section{Introduction}

In many practical domains, missing data is a major challenge that has a significant effect on data interpretation and analysis. 
Missing data can lead to biased estimates \citep{ayilara2019impact} and limit our ability to study and draw conclusions. In addition, most machine learning models (ML) can be applied to complete datasets only. Hence, techniques for dealing with missing data are necessary. A simple approach is \textit{omission}, in which observations with missing values are discarded from further analysis. However, in most cases, the proportion of missing data is not negligible. A retrospective electronical medical record (EMR) dataset is likely to have a large amount of missing values in patients' records. The omission would lead to a significant loss of valuable information, resulting in an unrepresentative sample and strong bias. In practice, data \textit{imputation} methods are used to deal with missing data by filling in the missing data with artificial values. Such methods offer powerful tools for addressing missing data in large datasets with complex data patterns.

Data imputation is usually done by inferring the missing values based on observed data. 
Imputation of EMR remains a major challenge, as it includes irregular individualized time-series data. Although various imputation methods are available \citep{batista2003analysis, mazumder2010spectral, van2011mice}, many of these are not well designed for clinical data, as they fail to account for inherent individual longitudinal patterns and clinical characteristics. In recent years, novel deep learning approaches were developed for data imputation \citep{che2018recurrent, yoon2018gain}. However, such approaches usually require massive amounts of data and not always practical.

We developed a new approach, Time-Dependent Iterative imputation (TDI), for imputing individualized time-series data. TDI addresses both multivariate and longitudinal data, by integrating forward-filling and Iterative Imputer, a version of MICE (Multivariate Imputation by Chained Equation, \citealp{van2011mice}). The integration employs a patient, variable, and observation-specific dynamic weighting strategy, based on the clinical patterns of the data including missing rates and measurement frequency. TDI offers a simple and practical solution, that utilizes the underlying metadata available, without requiring massive amounts of data. Experiments on real-world clinical datasets demonstrated that our model outperforms state-of-the-art imputation, in both producing better value estimates, and improving predictive performance. 

\vspace{-0.1cm}
\section{Problem Formulation}

Let $X \in \mathbb{R}^{N \times T \times D}$ be a time-series clinical dataset, where $N$ is the number of patients, $D$ is the number of covariates, and each patient’s trajectory is represented by discrete time points indexed as $t \in\{1, \dots, T\}$. We use the term \textit{observation} for the vector of the $D$ covariates of a subject at a particular time-point. The covariates can be either time-dependent (longitudinal) or time-independent (static). The number of time points (observations) available for patient $i$ is denoted by $t_i \leq T$. Hence, the longitudinal data of subject $i$ is represented by $X^i=(X_1^i,\dots,X_{t_i}^i)$, where $X_t^i$ is the observation of subject $i$ at time point $t$. The actual time when the $t$-th observation is obtained is denoted by $s_t^i$. $X_t^i$ may be incomplete. We define $M \in \{0,1\}^{N \times T \times D}$ as the mask matrix of $X$, where $m_{t,d}^i=1$ if $x_{t,d}^i$ is observed and zero otherwise (See Figure \ref{fig:dataframe}).

Our goal is to impute the missing values in $X$. The imputed dataset $\Tilde{X} \in \displaystyle \R^{N \times T \times D}$ is $\Tilde{X} = M \odot X + (1-M) \odot \hat{X} $, where $\hat{X}$ is an estimate of $X$ and $\odot$ is the element-wise product of matrices.

\vspace{-0.1cm}
\section{Time-Dependent Iterative imputation (TDI)}
\vspace{-0.1cm}
\subsection{Preliminaries}
\label{Preliminaries}
Our method relies on two established imputation methods that are applied to the data matrix $X$. In Forward-filling, each missing value of a patient's variable is imputed by using its last observed value. Let $\Tilde{X}_F$ be the resulting dataset. Notably, $\Tilde{X}_F$ could remain incomplete, due to missing values before the first measurement of a variable, or variables that were not recorded for a patient at all. Additionally, let $\Tilde{X}_I$ be the imputed dataset after applying the multivariate Iterative Imputer algorithm to $X$. Specifically, we applied \textit{scikit-learn}’s IterativeImputer \citep{pedregosa122011}, which was inspired by MICE \citep{van2011mice}. The Iterative Imputer uses a regression model to predict the missing values of each feature based on the other features, in a round-robin fashion. It is applied on the entire dataset $X$, independently of $\Tilde{X}_F$. 

 For each patient $i$, time point $t$, and variable $d$, we define the time passed since the last observation as $\Delta t_d^i  = s_t^i - s_{t-1, d}^i$, where $s_{t-1, d}^i$ is the time of the previous record of variable $d$ of patient $i$ ($\Delta t_d^i \geq 0$). $\Delta t_d^i = 0$ corresponds to the current time. $\Delta t_d^i = \infty$ if there is no past measurement. Additionally, let $r_t^i$ denote the fraction of available values of patient $i$ at time point $t$: $r_t^i = \frac{1}{D}\sum_{d=1}^D m_{t,d}^i$. Finally, denote the measurement frequency of variable $d$ by $f_d$. Namely, $f_d$ is the inverse value of the cohort average time (in hours) between measurements of variable $d$.
 
 \vspace{-0.1cm}
\subsection{The proposed model}

We propose a new time-dependent approach that considers both multivariate and longitudinal data in missing values imputation. At each time-point, the imputed value is the weighted sum of two imputed values estimated by the Iterative Imputer ($\Tilde{X}_I$) and forward-filling ($\Tilde{X}_F$). The integration employs a \textbf{patient}, \textbf{variable}, and \textbf{observation}-specific dynamic weighting strategy, based on the clinical patterns of the data. The imputed value for variable $d$ of patient $i$ at time point $t$ is:
$$ \Tilde{x}_{t,d}^i = w_{t,d}^i \cdot \Tilde{x}_{F_{t,d}}^i  + (1-w_{t,d}^i) \cdot \Tilde{x}_{I_{t,d}}^i $$
Where $w : \displaystyle \R ^+ \rightarrow [0,1]$ is the following weight function:
$$w_{t,d}^i = \frac{1}{1+ f_d \cdot r_t^i \cdot \Delta t_d^i}$$
In other words, $w(\Delta t_d^i)$ assigns weights to the imputation estimations as a function of three factors:

\textbf{(i) Time elapsed since past measurement ($\Delta t_d^i$)}: The key idea is that higher weights are assigned to Forward-filling with more recent observed values (lower $\Delta t$), in a similar manner to the inference by a caregiver in the clinical practice, when there are recent available values. In contrast, for large $\Delta t$, relying on the multivariate distributions using the observed data of other covariates is plausible. Importantly, while $\Tilde{X}_I$ is a complete dataset, $\Tilde{X}_F$ can still contain missing values (see Section \ref{Preliminaries}). Such values are imputed solely based on the Iterative Imputer algorithm.

\textbf{(ii) Variable sampling rates ($f_d$)}: For variables measured less often than others, Forward-filling is assigned more weight, covering a longer period back.

\textbf{(iii) The observational availability rates ($r_t^i$)}: The Iterative Imputer initializes missing values with a naive estimate (e.g., mean). In time points with high missing rates, the imputed values might be estimated mainly based on this initial strategy. Adding $r_t^i$ to the weight function penalizes the Iterative Imputer at time points with higher missing rates.

While we require that $w$ will be a decay function, we do not limit it to a specific function family. It can be either chosen a priori or treated as a hyper-parameter and selected via cross-validation. The weights should remain between $0$ and $1$ to preserve clinically plausible imputed values.

\vspace{-0.1cm}
\section{Evaluation}
\vspace{-0.1cm}
We empirically evaluated our method and six other simple and state-of-the-art imputation methods, on real-world clinical datasets (see Section \ref{Results:Datasets}). Section \ref{BenchmarkImputation} describes the benchmark imputation methods. We evaluated the methods using two approaches:

 \textbf{Masking:} This test compares imputed values to their known true values. Here, in addition to the originally missing data, we randomly mask a fraction of the available values of each variable (i.e., values $x_{t,d}^i$ such that $m_{t,d}^i=1$). Then, we impute the resulting masked dataset and compare the imputed values of the masked data to the ground truth values using different metrics (see Table  \ref{table:diffmetrics}).

\textbf{Prediction:} Data imputation does not always provide substantial improvement to predictive models. Hence, this test aims to assess the impact of the imputation on the quality of predicting a clinical outcome. We consider two different types of predictive tasks: (1) \textit{Baseline} prediction: A single prediction for each patient, based on baseline data obtained on admission or a few hours thereafter. (2) \textit{Longitudinal} predictions: Repeated predictions during hospitalization (e.g., at every time point $t$), using the patient's baseline and longitudinal data up to the current time point (Figure \ref{fig:predictiontypes}).

\vspace{-0.1cm}
\section{Results}
\vspace{-0.2cm}
\subsection{Datasets}
\vspace{-0.1cm}
\label{Results:Datasets}
\textbf{MIMIC-III:} A public dataset of de-identified clinical care data with over 40,000 patients admitted to the Beth Israel Deaconess Medical Center, Boston, between 2001 to 2012 \citep{johnson2016mimic}. We extracted 30 longitudinal features, including vital signs, basic metabolic panel (BMP), hematology panel, and coagulation panel (Table \ref{table:mimictoc}).

\textbf{COVID-19:} A dataset containing EMRs of 3,293 COVID-19 inpatients admitted to two hospitals (detailed hidden for double-blind review).

Preprocessing details can be found in Section \ref{data_preprocessing}.
\vspace{-0.2cm}
\subsection{Masking}
\begin{figure}[ht]
\captionsetup{font={small}, skip=0pt}
\begin{center}
\includegraphics[width=\textwidth]{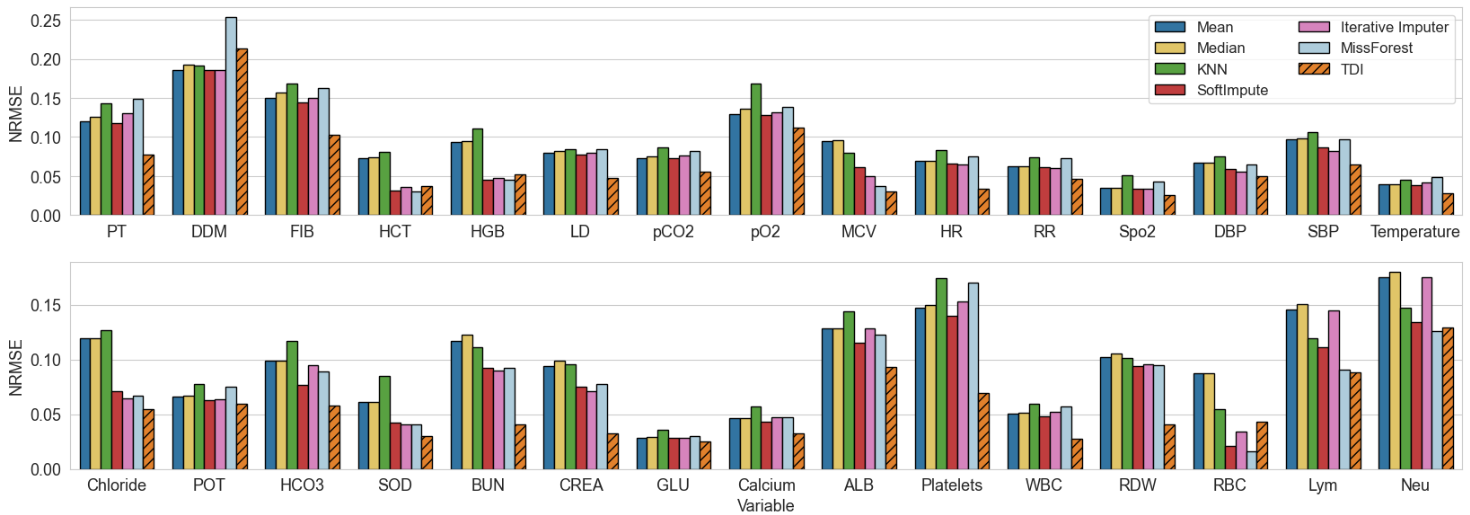}
\end{center}
\caption{\textbf{Masking performace.} A comparison of imputation models for each variable, using NRMSE (lower is better). TDI imputation (orange dashed bar) does best in 25 out of 30 variables.}
\label{fig:masking}
\end{figure}

\vspace{-0.1cm}
The masking experiment was conducted on a subsample of $559,837$ observations from $N=8000$ patients ($D=30$). Data standardization was performed, as required for some imputation methods (e.g., KNN). We randomly masked $10\%$ of the available values of each variable, and then imputed the data. We used three metrics to measure the difference between the imputed and real values (Table \ref{table:diffmetrics}). Figure \ref{fig:masking} summarizes the performance of seven imputation methods, in terms of normalized root mean squared error (NRMSE). TDI outperformed the other methods for 25 out of 30 variables. It also had the best overall score in all performance metrics, with an overall RMSE of 0.63, compared to 0.85 for SoftImpute, the second best method (see Table \ref{table:overallmasking}). Notably, after applying only forward-filling, our dataset still contains a large fraction of missing values (Table \ref{table:ffill_missingrate}). In a (possibly biased) evaluation tailored to test imputation by forward-filling, it was slightly inferior to TDI but outperformed the other methods (detailed in Section \ref{maskingffill}).

\vspace{-0.3cm}

\subsection{Prediction}
\begin{figure}[ht]
\captionsetup{font={small}, skip=0pt}
\begin{center}
\includegraphics[width=\textwidth]{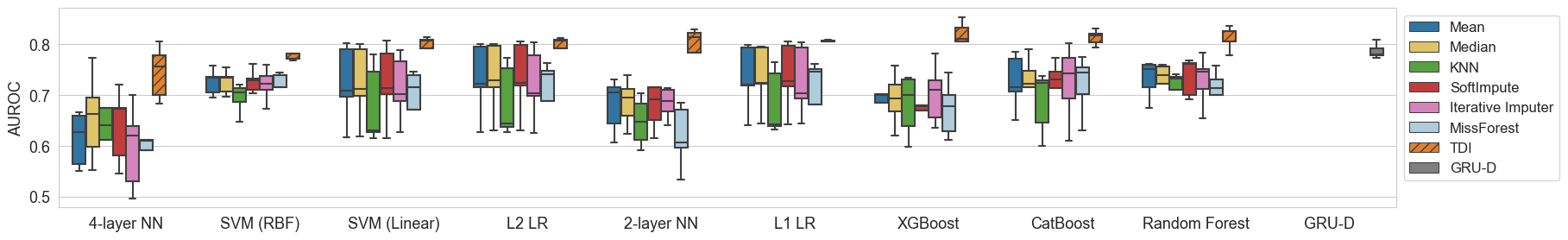}
\end{center}
\caption{\textbf{Performance of baseline mortality prediction in MIMIC-III, using 5-fold cross-validation}. TDI (orange dashed) improves the AUROC (y-axis) in all classifiers. Horizonal line: median.}
\label{fig:AUROCmimic2000}
\end{figure}

We performed a baseline binary classification task on MIMIC-III for predicting mortality in hospitalization or 30 days thereafter, based on data obtained in the first 48 hours after admission. We evaluated ten ML models for this prediction task, including feedforward neural network \citep{haykin1994neural}, SVM \citep{cortes1995support}, Logistic Regression, Random Forest \citep{breiman2001random}, CatBoost \citep{dorogush2018catboost}, XGBoost \citep{chen2016xgboost} and GRU-D \citep{che2018recurrent}. Note that GRU-D does not use an external imputation method. We measured the predictive performance in 5-fold cross-validation. In each iteration, an imputation method was first applied, and then each of the predictive models was trained on the imputed data. The imputation was done separately on the train and test folds. Each model's performance was measured using the area under the receiver-operator characteristics curve (AUROC) and the area under the precision-recall curve (AUPR) over the test folds. When applied to a subsample of N=2,000 patients, TDI improved the mean AUROC and AUPR in all classifiers, six of which perfomed better than GRU-D (Figures \ref{fig:AUROCmimic2000}, \ref{fig:AUPR_mimic_N=2000}). For a subsample of N=10,000 patients, GRU-D and CatBoost with TDI achieved comparable results (A difference of $0.1\%$ in favor of GRU-D in both metrics) (Table \ref{table:full_baseline_performance}). 
Similar results were achieved in a longitudinal setting on the COVID-19 datasets, for predicting mortality in the next 1-7 days (Supplementary \ref{fig:covid_prediction}). Section \ref{PredictionDesign} describes the experimental design of this benchmark.

\vspace{-0.3cm}
\section{Discussion}
\vspace{-0.2cm}
We presented TDI Imputation, a practical approach for missing data imputation designed for time-series clinical datasets. Our method imputes missing data by integrating forward-filling and the Iterative Imputer. The integration employs a patient, variable, and observation-specific dynamic weighting strategy, based on the clinical characteristics of the data. Compared to state-of-the-art imputation methods, our model achieved the best performance in the estimation of values of masked clinical variables with known ground truth. While being simple and applicable, TDI imputation led to an improved risk prediction in different ML models on real-world datasets. 
TDI can be readily applied multiple times with different seeds to account for imputation uncertainty (See \ref{MultipleImputation}).
Future work will consider additional prediction tasks on varying sample sizes, and compare to additional benchmark models. Extended experiments will test our method under various missingness mechanisms. We also intend to explore different weighting strategies.

\subsubsection*{Code and data availability}
The code for our method and for data preprocessing will be published in a GitHub repository upon publicity. The MIMIC-III dataset is available upon request at \href{http://dx.doi.org/10.13026/C2XW26}{http://dx.doi.org/10.13026/C2XW26}. Requests for access to the COVID-19 datasets should be directed to the hospitals.

\bibliography{iclr2023_conference}
\bibliographystyle{iclr2023_conference}

\newpage

\appendix
\section{Appendix}
\subsection{Benchmark imputation methods}
Our benchmark includes the following imputation methods:
\label{BenchmarkImputation}
\begin{itemize}
    \item \textit{Mean, Median}: The missing values of a certain variable are  replaced by the mean (or median) of the available ones.
    \item \textit{Forward-filling}: Each missing value of a patient’s variable is imputed by using its last observed value. Notably, the imputed dataset after using forward-filling often remains incomplete, due to missing values before the first measurement of a variable, or variables that were not recorded for a patient at all (Table \ref{table:ffill_missingrate}). A naive imputation of the remaining missing values with backward filling (i.e., using the next observed values) leads to severe leakage of future information and hence cannot be used in our framework. Therefore, forward-filling was evaluated in an adjusted masking experiment (detailed in Section \ref{maskingffill}) and excluded from the prediction experiment.
    \item \textit{SoftImpute} \citep{mazumder2010spectral}: Missing values are imputed using matrix completion by iterative soft thresholding of Singular Value Decomposition (SVD).
    \item \textit{KNN} \citep{batista2002study}: Missing values are imputed using the mean value of the \textit{k} nearest neighbors (closest samples) found in the training set.
    \item \textit{Iterative Imputer \citep{pedregosa122011}}: A version of the MICE algorithm \citep{van2011mice}. See \ref{Preliminaries} for additional information.
    \item \textit{MissForest} \citep{stekhoven2012missforest}: A non-parametric imputation method that uses random forests to predict the missing values, based on the observed values.
    \item \textit{GRU-D} \citep{che2018recurrent}: A deep learning model, based on Gated Recurrent Unit (GRU), that incorporates two representations of missing patterns, \textit{masking} and \textit{time interval}. Note that GRU-D \citep{che2018recurrent} is not explicitly designed for missing values imputation, and cannot be directly used in an unsupervised setting. Hence, it was used for prediction and excluded from the masking experiment.
\end{itemize}

\subsection{Datasets preprocessing details}
\label{data_preprocessing}

\subsubsection{MIMIC-III}
\label{mimic_preprocessing}
This section describes the preprocessing of the MIMIC-III dataset. MIMIC-III contains clinical care data for N=46,520 patients (patients available in both ADMISSIONS and PATIENTS data tables).

\textbf{Variable extraction and mapping.} We extracted 30 longitudinal features, including vital signs, basic metabolic panel (BMP), hematology panel, and coagulation panel (Supplementary Table  \ref{table:mimictoc}). Variables that were measured with different monitoring modalities were merged (for example, fingerstick glucose and blood glucose). We note that laboratory tests that were measured through other body fluids than blood (e.g., urine) were removed. Unit conversions were done when required. For the prediction tasks, we also utilized the patients' age and gender information. 

\textbf{Outlier removal.} To remove incorrect measurement values (due to typos for example), we manually defined with clinicians ranges of possible values for each longitudinal variable (including pathological ones). Values outside these ranges were excluded.

\textbf{Time discretization.} The temporal data was discretized into a 15-minute grid.

\textbf{Inclusion and exclusion criteria.} We included only patients aged between ages 18 and 89 (patients aged over 89 are masked in MIMIC) (N=36,564). In cases of several hospital admissions, we focused on the first admission of each patient. Only patients with at least one measurement of any of the longitudinal variables during the first hospitalization were included. The resulting cohort consists of 2,575,254 observations of N=35,968 patients, 

\subsubsection{COVID-19}
\label{covid_preprocessing}

We evaluated our proposed model on an additional retrospective cohort comprising two datasets from hospitals $A$ and $B$. The $A$ dataset consisted of all COVID-19 patients admitted to $A$ between March 2020 and March 2021 ($N_{A}=782$). The $B$ dataset consisted of all COVID-19 patients admitted to $B$ between March 2020 and April 2021 $N_{B}=2,511$). The study was reviewed and approved by the Institutional Review Boards (details removed for blinded refereeing).

The data used for this study included age, gender, and 13 longitudinal features, including laboratory test results and vital signs (See Supplementary Table \ref{table:covidtoc}). We mapped variables from the two hospitals and performed unit conversions when required. Similarly to the MIMIC-III preprocessing, we removed clinical outliers, namely, variable values that are out of predefined valid ranges. The temporal data was discretized to an hourly time grid, and multiple values of a test measured within the same hour were aggregated by mean. 

\subsection{Masking Evaluation with Forward Filling}
\label{maskingffill}
  Imputation by forward-filling is insufficient to enable prediction. After using forward-filling the dataset often has a large fraction of missing values (see Table \ref{table:ffill_missingrate}). Imputation of the remaining values cannot be done with Backward-filling, as it leads to severe leakage of future information. To properly compare forward-filling, we evaluated masking only on elements that had filled values by forward-filling, ignoring values left empty by the process. Note that this might bias our experiment towards highly measured variables and patients, a subset where it could be reasonable to use the most recent values with forward-filling. Figure \ref{fig:masking_ffill} summarizes the imputation performance on this subset, in terms of NRMSE. TDI outperformed the other methods for 21 out of 30 variables. It also had the best overall RMSE and NRMSE scores (Table \ref{table:ffill_overall_masking}). Forward filling demonstrated comparable results on this subset, with the best overall SMAPE score.

\subsection{Multiple imputation}
\label{MultipleImputation}
Single imputation does not account for the uncertainty in imputation estimates. Multiple imputation can be performed to measure uncertainty by providing valid standard errors and confidence intervals for the imputation estimates. That is, rather than generating a single imputed dataset $\Tilde{X}$ for a single data matrix $X$, we generate $m$ imputed datasets. These imputed datasets can then be utilized in the  subsequent pipeline (e.g., masking, prediction, and analysis) to derive multiple final results. This process can be used to better understand the variability in final results and to measure the uncertainty in estimation. As implemented in the IterativeImputer, our method can be readily used for multiple imputations by applying it repeatedly to the same dataset with different random seeds.

\subsection{Prediction tasks details}
\label{PredictionDesign}
\subsubsection{Baseline prediction}
We performed a baseline binary classification task on MIMIC-III for predicting mortality in hospitalization or 30 days thereafter, based on data obtained in the first 48 hours after admission. We evaluated ten ML models for this prediction task, including feedforward neural network \citep{haykin1994neural}, SVM \citep{cortes1995support}, Logistic Regression, Random Forest \citep{breiman2001random}, CatBoost \citep{dorogush2018catboost}, XGBoost \citep{chen2016xgboost} and GRU-D \citep{che2018recurrent}. Note that GRU-D doesn not use an external imputation method. While the considered classifiers cannot directly handle time series of different lengths for predicting a single target in time (e.g., mortality), the GRU-D uses a time-series sequence input for a single prediction. To perform a fair comparison, after data imputation, we sampled the time-series data to get a fixed-length input. Specifically, we used the two last observations in the first 48 hours for each patient. We also concatenated the masking vector along with the measurements, similarly as done in the GRU-D architecture. The GRU-D in contrast was fed initially with the full time-series dataset, giving it a potential advantage. To estimate the effect of the imputations on the predictive performance in different subsamples, we used 5-fold cross-validation. In each iteration, an imputation method was first applied, and then each of the predictive models was trained on the imputed data. The imputation was done separately on the train and test folds. Finally, the model performance was measured using the area under the receiver-operator characteristics curve (AUROC) and the area under the precision-recall curve (AUPR) over the test folds. 

To create the baseline prediction setting, exclusion/inclusion criteria were applied. The number of patients and observations remaining (out of N=2000 subsamples) after each criterion are listed:
\begin{itemize}
    \item Exclude patients who died less than 60 hours since admission (N=1,941 patients, 145,287 observations), as we wish to predict at least 12 hours in advance.
    \item Include observations from the first 48 hours (N=1,938 patients, 45,492 observations).
    \item Exclude empty observations, namely, such that all 30 longitudinal features are missing (N=1938 patients, 44,659 observations).
    \item Include patients with at least three time-points (N=1,740 patients, 44,312 observations).
\end{itemize}

The resulting cohort contains 1,740 patients, of which 156 patients had a positive outcome (8.9$\%$).

\subsubsection{Longitudinal predictions}
We performed a binary classification task for every hourly observation to predict mortality in one to seven days.  Patient observations where mortality was reported in the next seven days were called positive, and the rest were called negative. Observations from the 24-hours prior to the target event were excluded.

The initial cohort included 3,293 (195,237 observations) patients from both datasets.   
To create the longitudinal prediction setting, the following exclusion/inclusion criteria were applied: 
\begin{itemize}
    \item Exclude pregnant women and patients aged $\leq$ 18 (N=2,940 patients).
    \item Include only patients admitted for more than 24 hours (N=2,563 patients).
    \item Exclude patients who died less than 48 hours since admission (N=2,534 patients).
    \item Exclude observations recorded in the 24-h gap prior to the target, as we wish to predict in advance.
\end{itemize}

The resulting cohort consists of 43,812 hourly observations of 2,534 patients. 3,487 ($7.9\%$) observations of 453 patients were labeled positive.

\newpage
\subsection{Supplemental tables and figures}
\renewcommand\thefigure{A.\arabic{figure}}    
\setcounter{figure}{0}  

\begin{figure}[ht]
\centering
        \begin{subfigure}[b]{0.36\textwidth}
                \includegraphics[width=\linewidth]{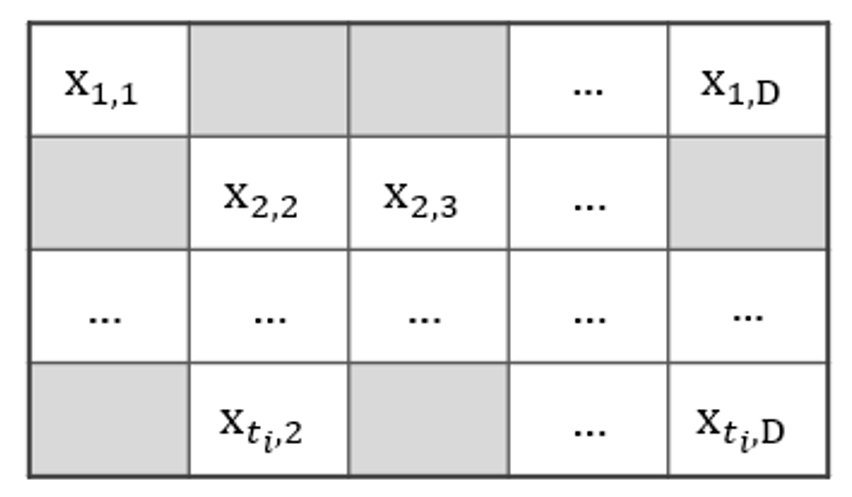}
                \caption{Data Matrix}
        \end{subfigure}%
        \begin{subfigure}[b]{0.36\textwidth}
                \includegraphics[width=\linewidth]{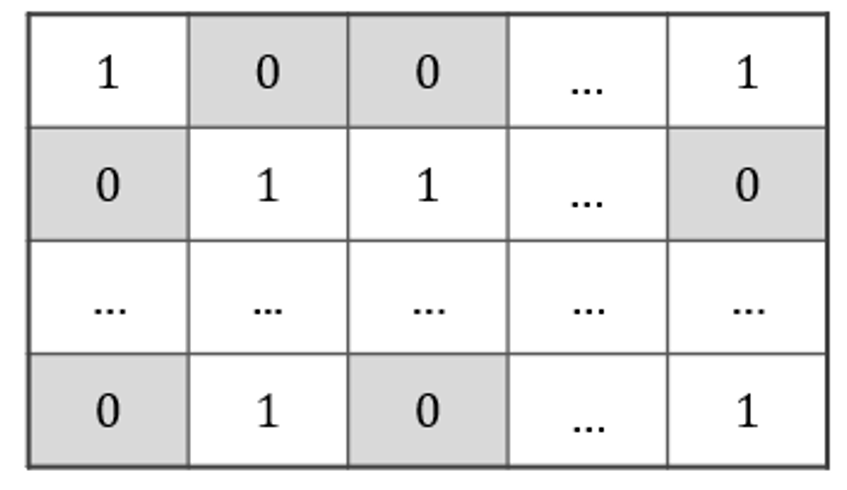}
                \caption{Mask Matrix}
        \end{subfigure}
        \caption{\textbf{Patient data}. (a) The longitudinal data matrix of patient $i$, denoted as $X^i$. Each patient's matrix consists of $D$ covariates (columns) and a different number of time-points $t_i$ (rows). Missing values are shaded. (b) A mask matrix that indicates which values of $X$ are observed.}
        \label{fig:dataframe}
\end{figure}

\vspace{2cm}

\begin{figure}[ht]
\centering
        \begin{subfigure}[b]{\textwidth}
                \includegraphics[width=\linewidth]{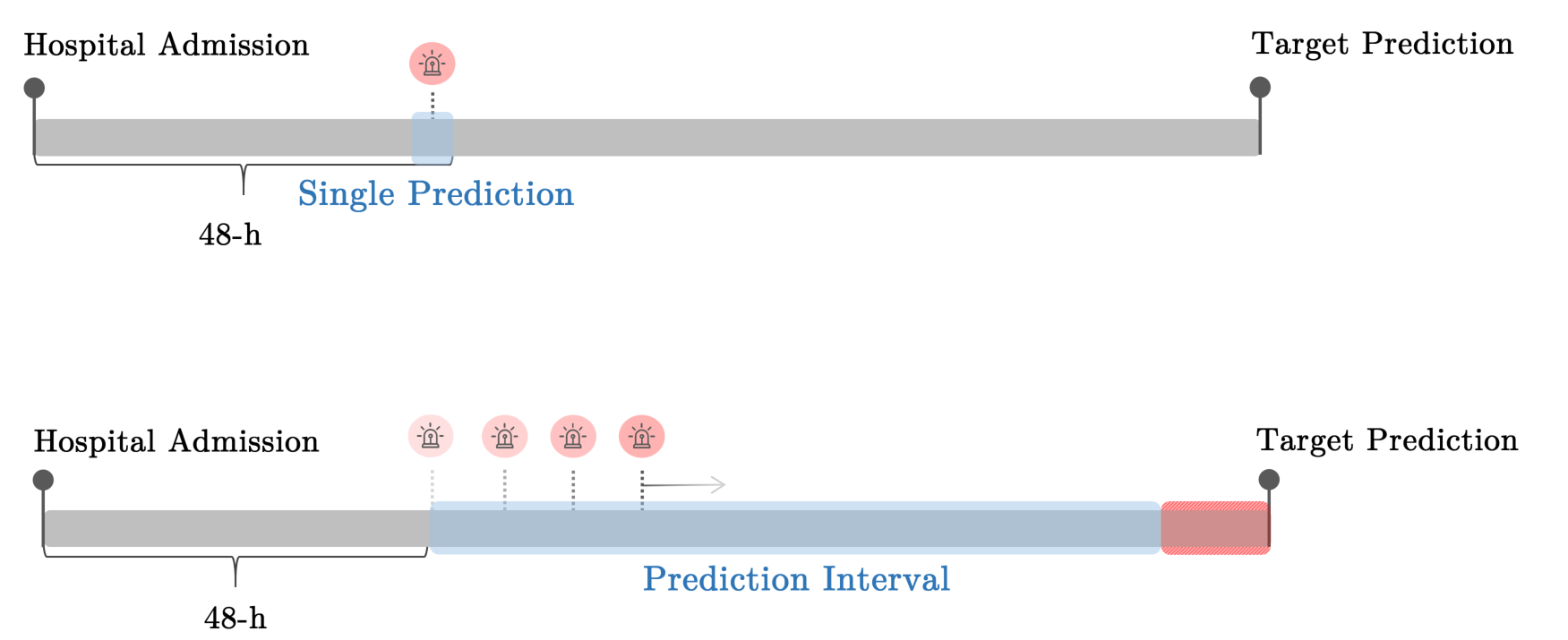}
        \end{subfigure}%
        \caption{\textbf{Predictive tasks along the patient timeline.} \textbf{Top: Baseline prediction.} The model generates a single prediction for each patient, based on baseline data from the first 48-h after admission. \textbf{Bottom: Longitudinal prediction.} The model generates longitudinal predictions for each patient, based on data from the entire hospitalization period. The blue areas refer to the time interval when the predictions are made. The Red dashed area  represents blocked prediction periods during which no predictions are made.}
        \label{fig:predictiontypes}
\end{figure}

\newpage

\begin{figure}[ht]
\centering
\begin{center}
\includegraphics[width=\textwidth]{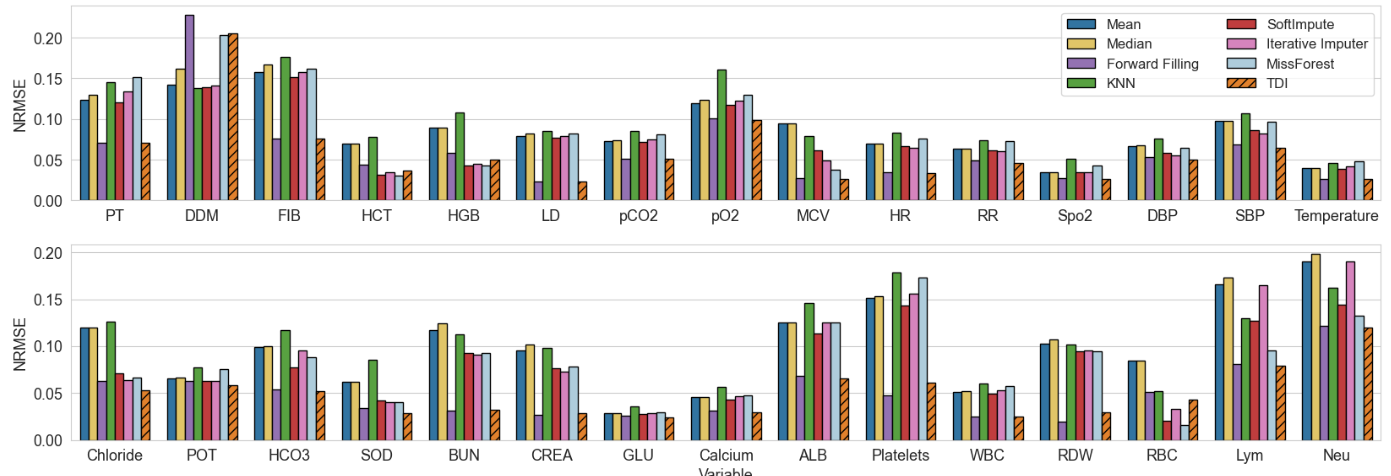}
\end{center}
\caption{\textbf{Masking performance on the forward-filled subset.} A comparison of imputation models for each variable, using NRMSE (lower is better). This evaluation was focused on elements that had filled values by forward-filling. TDI imputation (orange dashed bar) does best in 21 out of 30 variables.}
\label{fig:masking_ffill}
\end{figure}

\vspace{1cm}

\begin{figure}[ht]
\centering
        \begin{subfigure}[b]{\textwidth}
                \includegraphics[width=\linewidth]{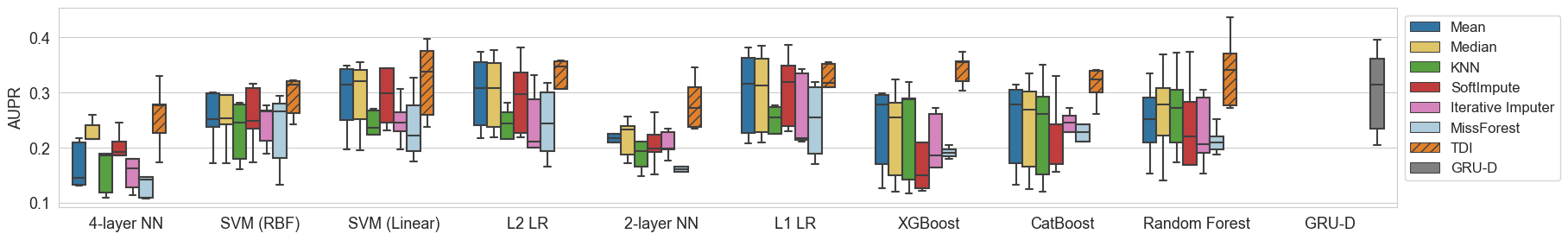}
        \end{subfigure}%
        \caption{\textbf{Performance of baseline mortality prediction in MIMIC-III, using 5-fold cross-validation}. TDI (orange dashed) improves the AUPR (y-axis) in all classifiers. Horizonal line: median.}
        \label{fig:AUPR_mimic_N=2000}
\end{figure}

\vspace{1cm}

\begin{figure}[ht]
\centering
        \begin{subfigure}[b]{\textwidth}
                \includegraphics[width=\linewidth]{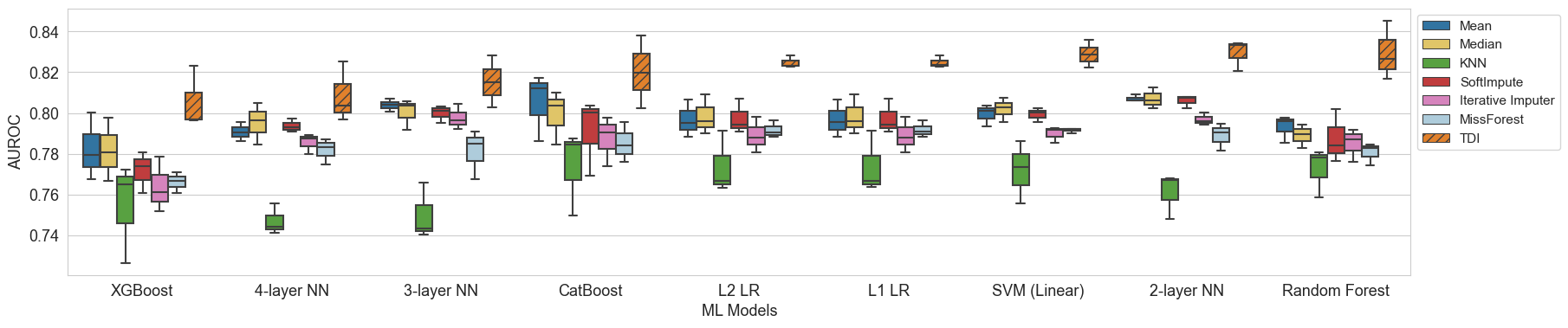}
        \end{subfigure}%
\\
        \begin{subfigure}[b]{\textwidth}
                \includegraphics[width=\linewidth]{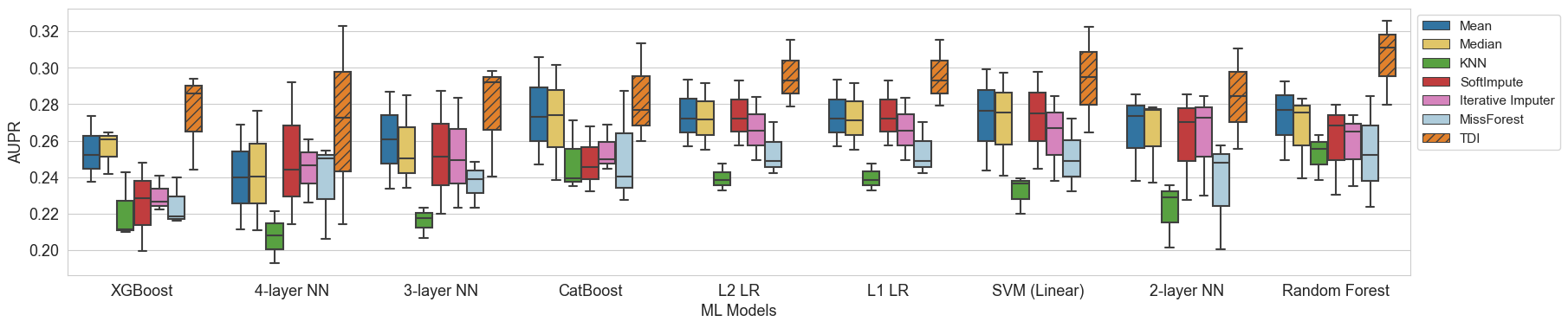}
        \end{subfigure}%
        \caption{\textbf{Performance of longitudinal mortality predictions on the COVID-19 data}. Comparison of machine learning models using 3-fold cross-validation. TDI (orange dashed) improves AUROC (top) and AUPR (bottom) in all classifiers. Horizonal line: median.}
        \label{fig:covid_prediction}
\end{figure}

\newpage

\makeatletter
\renewcommand\thetable{S\@arabic\c@table}
\makeatother

\begin{table}[!ht]
    \centering
    \begin{tabular}{|l|c|c|c|c|}
    \hline
        \textbf{Variable} & \textbf{Unit} & \textbf{N} & \textbf{Mean $\pm$ SD} & \textbf{Missing Rate (\%)} \\ \hline \hline
        \textbf{Heart Rate (HR)} & BPM & 1483322 & 86.73 $\pm$ 17.83 & 42.4 \\ \hline
        \textbf{Respiratory Rate (RR)} & BPM & 1482838 & 20.07 $\pm$ 5.83 & 42.42 \\ \hline
        \textbf{Spo2} & $\%$ & 1459840 & 96.93 $\pm$ 3.15 & 43.31 \\ \hline
        \textbf{Systolic Blood Pressure (SBP)} & mmHG & 1427439 & 121.47 $\pm$ 22.38 & 44.57 \\ \hline
        \textbf{Diastolic Blood Pressure (DBP)} & mmHG & 1426266 & 62.41 $\pm$ 14.55 & 44.62 \\ \hline
        \textbf{Glucose (GLU)} & mg/dL & 822380 & 136.43 $\pm$ 55.98 & 68.07 \\ \hline
        \textbf{Hematocrit (HCT)} & $\%$ & 464369 & 30.33 $\pm$ 4.88 & 81.97 \\ \hline
        \textbf{Potassium (POT)} & mEq/L & 443183 & 4.08 $\pm$ 0.59 & 82.79 \\ \hline
        \textbf{Hemoglobin (HGB)} & g/dL & 439220 & 10.3 $\pm$ 1.77 & 82.94 \\ \hline
        \textbf{Sodium (SOD)} & mEq/L & 422425 & 138.69 $\pm$ 5.03 & 83.6 \\ \hline
        \textbf{Temperature} & Celsius & 414649 & 36.98 $\pm$ 0.83 & 83.9 \\ \hline
        \textbf{Chloride} & mmol/L & 414014 & 104.05 $\pm$ 5.97 & 83.92 \\ \hline
        \textbf{Creatinine (CREA)} & mg/dL & 414084 & 1.41 $\pm$ 1.41 & 83.92 \\ \hline
        \textbf{Urea Nitrogen (BUN)} & mg/dL & 412401 & 28.61 $\pm$ 23.06 & 83.99 \\ \hline
        \textbf{Bicarbonate (HCO3)} & mEq/L & 404863 & 25.33 $\pm$ 4.76 & 84.28 \\ \hline
        \textbf{Platelet Count} & 10e3/$\mu$L & 396073 & 233.38 $\pm$ 148.75 & 84.62 \\ \hline
        \textbf{White Blood Cells (WBC)} & 10e3/$\mu$L & 379718 & 11.37 $\pm$ 6.93 & 85.26 \\ \hline
        \textbf{Red Blood Cells (RBC)} & m/$\mu$L & 377993 & 3.42 $\pm$ 0.6 & 85.32 \\ \hline
        \textbf{MCV} & fL & 377547 & 89.86 $\pm$ 6.2 & 85.34 \\ \hline
        \textbf{Red Cell Distribution Width (RDW)} & $\%$ & 377251 & 15.54 $\pm$ 2.32 & 85.35 \\ \hline
        \textbf{pCO2} & mmHg & 349334 & 41.85 $\pm$ 10.11 & 86.43 \\ \hline
        \textbf{pO2} & mmHg & 349387 & 141.19 $\pm$ 90.66 & 86.43 \\ \hline
        \textbf{Calcium} & mg/dL & 326887 & 8.39 $\pm$ 0.8 & 87.31 \\ \hline
        \textbf{PT} & Sec & 248814 & 16.02 $\pm$ 4.97 & 90.34 \\ \hline
        \textbf{Albumin (ALB)} & g/dL & 63613 & 2.9 $\pm$ 0.67 & 97.53 \\ \hline
        \textbf{Lactate Dehydrogenase (LD)} & IU/L & 54444 & 530.19 $\pm$ 1045.02 & 97.89 \\ \hline
        \textbf{Lymphocytes (Lym)} & $\%$ & 49035 & 15.57 $\pm$ 16.88 & 98.1 \\ \hline
        \textbf{Neutrophils (Neu)} & $\%$ & 48818 & 74.39 $\pm$ 18.88 & 98.1 \\ \hline
        \textbf{Fibrinogen (FIB)} & mg/dL & 29644 & 312.39 $\pm$ 192.28 & 98.85 \\ \hline
        \textbf{D-Dimer (DDM)} & ng/mL & 1826 & 4128.61 $\pm$ 3435.9 & 99.93 \\ \hline
    \end{tabular}
    \caption{List of 30 variables after preprocessing the MIMIC-III dataset.}
    \label{table:mimictoc}
\end{table}

\begin{table}[!ht]
    \centering
    \begin{tabular}{|l|c|c|c|c|}
    \hline
        \textbf{Variable} & \textbf{Unit} & \textbf{N} & \textbf{Mean ± SD} & \textbf{Missing Rate (\%)} \\ \hline
        \textbf{Heart Rate (HR)} & BPM & 129026 & 85.88 ± 18.94 & 33.91 \\ \hline
        \textbf{Temperature} & Celsius & 80513 & 36.9 ± 0.82 & 58.76 \\ \hline
        \textbf{SBP} & mmHG & 70818 & 128.46 ± 22.36 & 63.73 \\ \hline
        \textbf{DBP} & mmHG & 70564 & 69.97 ± 14.55 & 63.86 \\ \hline
        \textbf{Saturation} & \% & 68386 & 94.77 ± 5.04 & 64.97 \\ \hline
        \textbf{Glucose} & mg/dL & 51713 & 164.43 ± 76.8 & 73.51 \\ \hline
        \textbf{HCO3} & mmol/L & 31083 & 29.42 ± 7.11 & 84.08 \\ \hline
        \textbf{Platelets} & 10e3/$\mu$L & 18658 & 244.54 ± 126.22 & 90.44 \\ \hline
        \textbf{BUN} & mg/dL & 18473 & 32.34 ± 26.56 & 90.54 \\ \hline
        \textbf{Neutrophils (\#)} & 10e3/$\mu$L & 18431 & 8.12 ± 5.53 & 90.56 \\ \hline
        \textbf{Lymphocytes  (\#)} & 10e3/$\mu$L & 18261 & 1.25 ± 1.29 & 90.65 \\ \hline
        \textbf{Albumin} & g/L & 12755 & 31.46 ± 6.84 & 93.47 \\ \hline
        \textbf{LDH} & U/L & 11416 & 844.54 ± 2175.36 & 94.15 \\ \hline
    \end{tabular}
        \caption{List of 13 variables after preprocessing the COVID-19 datasets.}
    \label{table:covidtoc}

\end{table}

\newpage

\begin{table}[!ht]
    \centering
    \begin{tabular}{|l|c|}
    \hline
        \textbf{Evaluation Metric} & \textbf{Formula} \\ \hline \hline
         &  \\
        RMSE & $\sqrt{\frac{1}{n}\sum_{i=1}^{n}{\Big(y_i - \hat{y_i}}\Big)^2}$ \\
         &  \\
        NRMSE & $\frac{\text{RMSE}
}{y_{max} - y_{min}}$ \\ 
        &  \\
        SMAPE & $\frac{1}{n}\Sigma_{i=1}^{n} \frac{|\hat{y_i} - y_i|}{(\hat{y_i} + y_i) / 2}$   \\
        & \\ \hline
    \end{tabular}
    \caption{\textbf{Evaluation measures of differences between values.} $y$ and $\hat{y}$ refer to the actual and estimated values, respectively. RMSE: Root Mean Squared Error. NRMSE: Normalized Root Mean. SMAPE: Symmetric mean absolute percentage error}
    \label{table:diffmetrics}
\end{table}

\vspace{2cm}

\begin{table}[!ht]
    \centering
    \begin{tabular}{|l|c|c|c|}
    \hline
        \textbf{Imputation} & \textbf{RMSE} & \textbf{NRMSE} & \textbf{SMAPE} \\ \hline \hline
        \textbf{TDI} & 0.636 & 0.060 & 0.858 \\ 
        \textbf{SoftImpute} & 0.851 & 0.079 & 1.365 \\ 
        \textbf{Iterative Imputer} & 0.891 & 0.084 & 1.416 \\ 
        \textbf{MissForest} & 0.932 & 0.086 & 1.058 \\ 
        \textbf{Mean} & 1.013 & 0.095 & 1.985 \\ 
        \textbf{Median} & 1.031 & 0.097 & 1.446 \\ 
        \textbf{KNN} & 1.124 & 0.102 & 1.324 \\ \hline
    \end{tabular}
    \caption{\textbf{Masking overall performance.} The average value across all variables in MIMIC-III, for difference evaluation metrics.}
    \label{table:overallmasking}
\end{table}

\begin{table}[!ht]
    \centering
    \begin{tabular}{|l|c|}
    \hline
        \textbf{} \textbf{Variable} & \textbf{Missing Rate (\%)} \\ \hline
        \hline
        \textbf{Glucose (GLU)} & 2.08 \\ \hline
        \textbf{Hemoglobin (HGB)} & 3.68 \\ \hline
        \textbf{Hematocrit (HCT)} & 4.13 \\ \hline
        \textbf{Creatinine (CREA)} & 4.45 \\ \hline
        \textbf{Urea Nitrogen (BUN)} & 4.45 \\ \hline
        \textbf{Platelet Count} & 4.55 \\ \hline
        \textbf{Chloride} & 4.56 \\ \hline
        \textbf{Bicarbonate (HCO3)} & 4.59 \\ \hline
        \textbf{White Blood Cells (WBC)} & 4.93 \\ \hline
        \textbf{Red Blood Cells (RBC)} & 4.94 \\ \hline
        \textbf{Red Cell Distribution Width (RDW)} & 4.96 \\ \hline
        \textbf{Potassium (POT)} & 4.98 \\ \hline
        \textbf{MCV} & 4.98 \\ \hline
        \textbf{Sodium (SOD)} & 5.21 \\ \hline
        \textbf{Calcium} & 9.6 \\ \hline
        \textbf{PT} & 10.23 \\ \hline
        \textbf{pO2} & 20.2 \\ \hline
        \textbf{pCO2} & 20.2 \\ \hline
        \textbf{Respiratory Rate (RR)} & 26.85 \\ \hline
        \textbf{Heart Rate (HR)} & 26.89 \\ \hline
        \textbf{Spo2} & 26.97 \\ \hline
        \textbf{Systolic Blood Pressure (SBP)} & 27.05 \\ \hline
        \textbf{Diastolic Blood Pressure (DBP)} & 27.05 \\ \hline
        \textbf{Temperature} & 29.0 \\ \hline
        \textbf{Albumin (ALB)} & 40.69 \\ \hline
        \textbf{Neutrophils (Neu)} & 47.62 \\ \hline
        \textbf{Lymphocytes (Lym)} & 47.75 \\ \hline
        \textbf{Lactate Dehydrogenase (LD)} & 49.44 \\ \hline
        \textbf{Fibrinogen (FIB)} & 63.87 \\ \hline
        \textbf{D-Dimer (DDM)} & 94.79 \\ \hline
    \end{tabular}
    \caption{\textbf{Missing rates after forward-filling imputation}. The variable missing rates (observation-wise) after using forward-filling imputation on the MIMIC-III masking sample (N=8,000).}
\label{table:ffill_missingrate}
\end{table}

\begin{table}[!ht]
    \centering
    \begin{tabular}{|l|c|c|c|}
    \hline
        \textbf{Imputation} & \textbf{RMSE} & \textbf{NRMSE} & \textbf{SMAPE} \\ \hline
        \hline
        \textbf{TDI} & 0.573 & 0.054 & 0.736 \\ \hline
        \textbf{Forward Filling} & 0.601 & 0.056 & 0.717 \\ \hline
        \textbf{SoftImpute} & 0.846 & 0.078 & 1.366 \\ \hline
        \textbf{Iterative Imputer} & 0.886 & 0.083 & 1.417 \\ \hline
        \textbf{MissForest} & 0.92 & 0.084 & 1.059 \\ \hline
        \textbf{Mean} & 1.007 & 0.094 & 1.984 \\ \hline
        \textbf{Median} & 1.029 & 0.097 & 1.459 \\ \hline
        \textbf{KNN} & 1.118 & 0.101 & 1.328 \\ \hline
    \end{tabular}
    \caption{\textbf{Masking performance on the forward-filled subset}. The average value across all variables in MIMIC-III, for difference evaluation metrics.}
\label{table:ffill_overall_masking}
\end{table}

\begin{table}[]
\begin{tabular}{|c|l|l|ccc|ccc|}
\hline
\textbf{}                                             & \textbf{}              & \textbf{}           & \multicolumn{3}{c|}{\textbf{AUROC}}                                                      & \multicolumn{3}{c|}{\textbf{AUPR}}                                                       \\ \hline
\textbf{N}                                            & \textbf{ML Model}      & \textbf{Imputation} & \multicolumn{1}{l|}{\textbf{Mean}}  & \multicolumn{1}{l|}{\textbf{Median}} & \textbf{SD} & \multicolumn{1}{l|}{\textbf{Mean}}  & \multicolumn{1}{l|}{\textbf{Median}} & \textbf{SD} \\ \hline \hline
\multicolumn{1}{|c|}{{\textbf{2,000}}} & \textbf{Random Forest} & TDI                 & \multicolumn{1}{l|}{\textbf{0.814}} & \multicolumn{1}{l|}{\textbf{0.825}}  & 0.023       & \multicolumn{1}{l|}{\textbf{0.339}} & \multicolumn{1}{l|}{\textbf{0.342}}  & 0.069       \\ \cline{2-9} 
\multicolumn{1}{|c|}{}                                & \textbf{CatBoost}      & TDI                 & \multicolumn{1}{l|}{0.813}          & \multicolumn{1}{l|}{0.817}           & 0.015       & \multicolumn{1}{l|}{0.313}          & \multicolumn{1}{l|}{0.323}           & 0.033       \\ \cline{2-9} 
\multicolumn{1}{|c|}{}                                & \textbf{XGBoost}       & TDI                 & \multicolumn{1}{l|}{0.809}          & \multicolumn{1}{l|}{0.81}            & 0.042       & \multicolumn{1}{l|}{0.342}          & \multicolumn{1}{l|}{0.355}           & 0.029       \\ \cline{2-9} 
\multicolumn{1}{|c|}{}                                & \textbf{L1 LR}         & TDI                 & \multicolumn{1}{l|}{0.792}          & \multicolumn{1}{l|}{0.807}           & 0.033       & \multicolumn{1}{l|}{0.313}          & \multicolumn{1}{l|}{0.317}           & 0.05        \\ \cline{2-9} 
\multicolumn{1}{|c|}{}                                & \textbf{2-layer NN}    & TDI                 & \multicolumn{1}{l|}{0.791}          & \multicolumn{1}{l|}{0.814}           & 0.051       & \multicolumn{1}{l|}{0.28}           & \multicolumn{1}{l|}{0.272}           & 0.048       \\ \cline{2-9} 
\multicolumn{1}{|c|}{}                                & \textbf{L2 LR}         & TDI                 & \multicolumn{1}{l|}{0.791}          & \multicolumn{1}{l|}{0.806}           & 0.033       & \multicolumn{1}{l|}{0.316}          & \multicolumn{1}{l|}{0.347}           & 0.062       \\ \cline{2-9} 
\multicolumn{1}{|c|}{}                                & \textbf{GRU-D}         & GRU-D               & \multicolumn{1}{l|}{0.786}          & \multicolumn{1}{l|}{0.78}            & 0.015       & \multicolumn{1}{l|}{0.302}          & \multicolumn{1}{l|}{0.314}           & 0.082       \\ \cline{2-9} 
\multicolumn{1}{|c|}{}                                & \textbf{SVM (Linear)}  & TDI                 & \multicolumn{1}{l|}{0.786}          & \multicolumn{1}{l|}{0.807}           & 0.045       & \multicolumn{1}{l|}{0.322}          & \multicolumn{1}{l|}{0.338}           & 0.071       \\ \cline{2-9} 
\multicolumn{1}{|c|}{}                                & \textbf{SVM (RBF)}     & TDI                 & \multicolumn{1}{l|}{0.785}          & \multicolumn{1}{l|}{0.772}           & 0.027       & \multicolumn{1}{l|}{0.292}          & \multicolumn{1}{l|}{0.314}           & 0.037       \\ \cline{2-9} 
\multicolumn{1}{|c|}{}                                & \textbf{4-layer NN}    & TDI                 & \multicolumn{1}{l|}{0.745}          & \multicolumn{1}{l|}{0.756}           & 0.052       & \multicolumn{1}{l|}{0.257}          & \multicolumn{1}{l|}{0.277}           & 0.06        \\ \hline \hline
{\textbf{10,000}}                      & \textbf{GRU-D}         & GRU-D               & \multicolumn{1}{l|}{\textbf{0.817}} & \multicolumn{1}{l|}{\textbf{0.818}}  & 0.016       & \multicolumn{1}{l|}{\textbf{0.363}} & \multicolumn{1}{l|}{0.359}           & 0.041       \\ \cline{2-9} 
& \textbf{CatBoost}      & TDI                 & \multicolumn{1}{l|}{0.816}          & \multicolumn{1}{l|}{0.817}           & 0.012       & \multicolumn{1}{l|}{0.362}          & \multicolumn{1}{l|}{\textbf{0.365}}  & 0.022       \\ \cline{2-9} 
& \textbf{L1 LR}         & TDI                 & \multicolumn{1}{l|}{0.8}            & \multicolumn{1}{l|}{0.801}           & 0.005       & \multicolumn{1}{l|}{0.344}          & \multicolumn{1}{l|}{0.359}           & 0.032       \\ \cline{2-9} 
& \textbf{L2 LR}         & TDI                 & \multicolumn{1}{l|}{0.799}          & \multicolumn{1}{l|}{0.799}           & 0.005       & \multicolumn{1}{l|}{0.342}          & \multicolumn{1}{l|}{0.358}           & 0.034       \\ \cline{2-9} 
& \textbf{SVM (Linear)}  & TDI                 & \multicolumn{1}{l|}{0.798}          & \multicolumn{1}{l|}{0.797}           & 0.004       & \multicolumn{1}{l|}{0.338}          & \multicolumn{1}{l|}{0.354}           & 0.039       \\ \cline{2-9} 
& \textbf{Random Forest} & TDI                 & \multicolumn{1}{l|}{0.797}          & \multicolumn{1}{l|}{0.797}           & 0.012       & \multicolumn{1}{l|}{0.344}          & \multicolumn{1}{l|}{0.346}           & 0.017       \\ \cline{2-9} 
& \textbf{XGBoost}       & TDI                 & \multicolumn{1}{l|}{0.794}          & \multicolumn{1}{l|}{0.79}            & 0.01        & \multicolumn{1}{l|}{0.322}          & \multicolumn{1}{l|}{0.332}           & 0.025       \\ \cline{2-9} 
& \textbf{SVM (RBF)}     & TDI                 & \multicolumn{1}{l|}{0.762}          & \multicolumn{1}{l|}{0.76}            & 0.008       & \multicolumn{1}{l|}{0.323}          & \multicolumn{1}{l|}{0.328}           & 0.022       \\ \cline{2-9} 
& \textbf{2-layer NN}    & TDI                 & \multicolumn{1}{l|}{0.748}          & \multicolumn{1}{l|}{0.747}           & 0.004       & \multicolumn{1}{l|}{0.283}          & \multicolumn{1}{l|}{0.292}           & 0.023       \\ \cline{2-9} 
& \textbf{4-layer NN}    & TDI                 & \multicolumn{1}{l|}{0.688}          & \multicolumn{1}{l|}{0.695}           & 0.014       & \multicolumn{1}{l|}{0.251}          & \multicolumn{1}{l|}{0.252}           & 0.025       \\ \hline
\end{tabular}
\caption{\textbf{Performance of baseline mortality prediction in MIMIC-III, using 5-fold cross-validation.} For each classifier, only the best performer imputation method is presented. GRU-D does not use external imputation methods. Best measures per sample size are bolded.}
\label{table:full_baseline_performance}
\end{table}

\end{document}